\documentclass{article} 
\usepackage{iclr2025,times}

\usepackage{amsmath,amsfonts,bm}

\def\eqref#1{equation~\ref{#1}}

\def\1{\bm{1}}

\DeclareMathAlphabet{\mathsfit}{\encodingdefault}{\sfdefault}{m}{sl}
\SetMathAlphabet{\mathsfit}{bold}{\encodingdefault}{\sfdefault}{bx}{n}

\usepackage{graphicx}
\usepackage{hyperref}
\usepackage{url}
\usepackage{multirow}
\usepackage{booktabs}
\usepackage{subfigure}

\title{From Fog to Failure: The Unintended Consequences of Dehazing on Object Detection in Clear Images}

\author{Ashutosh Kumar \\
School of Information\\
Rochester Institute of Technology\\
Rochester, NY 14623, USA \\
\texttt{ak1825@rit.edu} \\
\And
Aman Chadha\thanks{Work done outside position at Amazon.} \\
Amazon Gen AI \\
Santa Clara, CA, USA \\
\texttt{hi@aman.ai} \\
}

\iclrfinalcopy 
\begin{document}

\maketitle

\begin{abstract}
This study explores the challenges of integrating human visual cue-based dehazing into object detection, given the selective nature of human perception. While human vision adapts dynamically to environmental conditions, computational dehazing does not always enhance detection uniformly. We propose a multi-stage framework where a lightweight detector identifies regions of interest (RoIs), which are then improved via spatial attention-based dehazing before final detection by a heavier model. Though effective in foggy conditions, this approach unexpectedly degrades the performance on clear images. We analyze this phenomenon, investigate possible causes, and offer insights for designing hybrid pipelines that balance enhancement and detection. Our findings highlight the need for selective preprocessing and challenge assumptions about universal benefits from cascading transformations. The implementation of the framework is available here\footnote{\href{https://github.com/ashu1069/perceptual-piercing}{https://github.com/ashu1069/perceptual-piercing}}.
\end{abstract}
\section{Introduction}
Low-visibility conditions, such as rain, snow, fog, smoke, and haze, pose significant challenges for deep learning applications in autonomous vehicles, security and surveillance, maritime navigation, and agricultural robotics. Object detection models struggle in these environments due to reduced contrast and obscured features, often leading to performance degradation. While image enhancement methods, including dehazing, improve visibility, they can also introduce artifacts or distortions that negatively impact downstream tasks. Overprocessing may lead to false positives and increased computational overhead, highlighting the need for more selective enhancement strategies. Motivated by real-world challenges, such as disruptions in airport operations where poor visibility delays taxiing and docking, this study proposes a vision-inspired deep learning framework tailored for adverse conditions, particularly fog.

To address these challenges, we introduce \textbf{Selective Region Enhancement}, a method that focuses on specific regions of interest rather than applying uniform dehazing. This approach reduces processing overhead and prevents unintended degradations that may introduce false positives. Additionally, we propose \textbf{Integration with Object Detection}, bridging image enhancement with object detection in a unified pipeline. This integration leverages the strengths of both techniques, overcoming limitations of traditional independent processing models. Our approach draws inspiration from human visual mechanisms, including selective attention, foveal and peripheral vision, adaptive eye responses, bottom-up sensory cues, and top-down goal-driven processing (see Appendix~\ref{appendix:human_visual_cues}).

The paper is structured as follows: Section \ref{sec:related_work} reviews prior work on low-visibility object detection and integration of human visual cues in deep learning applications. Section \ref{sec:methodology} presents our framework, including its vision-inspired design, data set selection, and experimental setup, followed by results and observed anomalies in Section~\ref{sec:results}.
\section{Related Work}
\label{sec:related_work}
Advancements in navigation and detection under low-visibility conditions have leveraged sensor fusion, visual cue integration, and computational techniques. Aircraft landing studies have explored sensor fusion of visible and virtual imagery \citep{b1} and visual-inertial navigation using runway features \citep{b2}. Multi-sensor fusion algorithms have improved odometry in GPS-denied environments \citep{b3}, while research on depth visualization has enhanced navigation and obstacle avoidance \citep{b4}. Synthetic Vision Systems and full-windshield Head-Up Displays aid drivers and pilots in low visibility \citep{b5,b8}. Image enhancement techniques for low-light conditions \citep{b6} and the fusion of visual cues with wireless communication improve road safety \citep{b7}. Studies have also emphasized the role of geometrical shapes and colors in driving perception via Head-Up Displays \citep{b9}.

Despite these advances, challenges persist, including computational complexity \citep{b2,b6,b10}, performance issues under extreme conditions \citep{b3,b7}, overfitting due to limited datasets \citep{b2,b3}, and insufficient real-world validation \citep{b1,b7,b10}. Some works lack rigorous validation \citep{b5,b9}.

In visual recognition, research has explored human-like processing in computational models. Studies on brain mechanisms highlight hierarchical, feedforward object recognition \citep{b11}, while comparisons with deep neural networks (DNNs) reveal human superiority in handling distortions and attention mechanisms \citep{b12,b13}. Eye-tracking data has been used to guide DNN attention with limited success \citep{b14}. Approaches such as adversarial learning for feature discrimination \citep{b15}, biologically inspired top-down and bottom-up models \citep{b16}, and retina-mimicking models for dehazing \citep{b17} have been proposed. Foveal-peripheral dynamics have also been explored to balance computational efficiency and high-resolution perception \citep{b18}.

Recent research has tackled low-visibility challenges like fog, low light, and sandstorms. The YOLOv5s FMG algorithm improves small-target detection with enhanced modules \citep{b21}, while novel MLP-based networks refine image clarity in hazy and sandstorm conditions \citep{b22}. The PKAL approach integrates adversarial learning and feature priors for robust recognition \citep{b23}. Deformable convolutions and attention mechanisms enhance pedestrian and vehicle detection in poor visibility \citep{b24}. Reviews highlight the limitations of non-learning and meta-heuristic dehazing methods in real-time applications \citep{b25}, emphasizing the need for integrated low-level and high-level vision techniques \citep{b26}. Innovations such as spatiotemporal attention for video sequences \citep{b27}, the PDE framework for simultaneous detection and enhancement \citep{b28}, spatial priors for saliency detection \citep{b29}, and early visual cues for object boundary detection \citep{b30} further contribute to the field.

Despite advances, existing methods struggle with joint optimization of object detection and image enhancement, detection of low-contrast objects, and adaptation to dynamic visibility changes. This paper addresses these challenges by integrating human visual cues, such as attention mechanisms and contextual understanding, into object detection, enhancing both robustness and efficiency. Traditional approaches process entire images uniformly, increasing computational load, and sometimes degrading clear regions. Our method selectively enhances regions of interest, reducing unnecessary computations and improving responsiveness under varying conditions.
\section{Methodology}
\label{sec:methodology}
\begin{figure}
    \centering
    \includegraphics[width=0.88\linewidth]{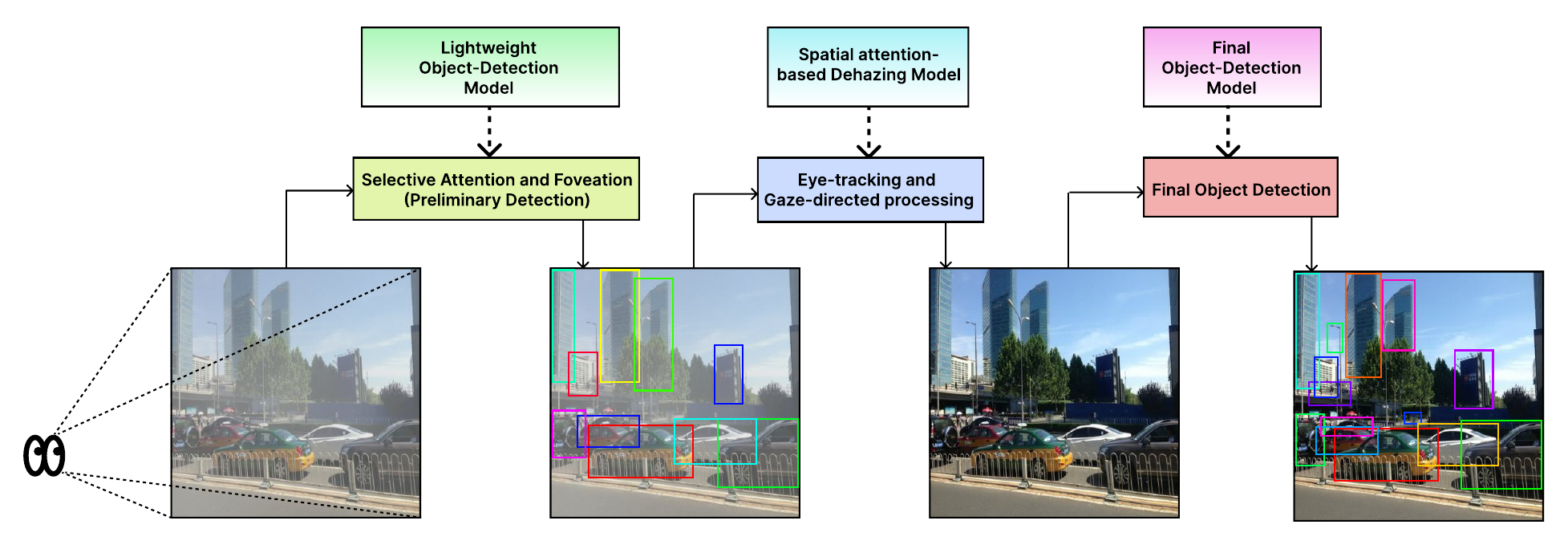}
    \caption{Overall architecture of Perceptual Piercing: (a) Preliminary detection using lightweight object detection model (b) Gaze-directed dehazing using spatial attention on region of interests (c) Final detection using a large and robust model.}
    \label{fig:framework}
\end{figure}

The proposed methodology, illustrated in Figure~\ref{fig:framework}, presents a deep learning framework that enhances object detection in low-visibility conditions by leveraging the atmospheric scattering model and human visual cortex principles. It integrates adaptive image enhancement with object detection, optimizing performance through different integration strategies. The pipeline starts with a lightweight detection model to identify regions of interest, guiding spatial attention in the dehazing process. This targeted enhancement preserves critical features while reducing computational overhead. A more robust detection model then refines and improves object recognition.

For dehazing, we train and evaluate state-of-the-art models, including AOD-Net\citep{li2017aod}, UNet-Dehaze\citep{zhou2024dehaze}, and DehazeNet\citep{cai2016dehazenet}, using the Foggy Cityscapes dataset~\citep{b31} (see Appendix\ref{appendix:datasets}). Table~\ref{tab:dehzing_methods} presents a comparative analysis of their dehazing performance, while Figure~\ref{fig:map_compare} illustrates their impact on object detection. Among these methods, AOD-Net demonstrated the best dehazing performance, prompting further architectural enhancements. This resulted in AOD-NetX, a spatial attention-enhanced version trained on the Foggy Cityscapes dataset. Detailed modifications to AOD-NetX are provided in Appendix~\ref{appendix:aodnetX} and Figure~\ref{fig:aodnetx}. For object detection models, we have used pre-trained YOLOv5 and YOLOv8 (see Appendix~\ref{appendix:object_detection}).

To ensure robust evaluation, we tested our pipeline on three datasets: Foggy Cityscapes, RESIDE-\(\beta\), and RESIDE-OTS (see Appendix~\ref{appendix:datasets})~\citep{b32}. This comprehensive assessment highlights the adaptability and effectiveness of our approach across multiple low-visibility scenarios.
\section{Results and Observed Anomalies}
\label{sec:results}
For in-distribution performance on the Foggy Cityscapes dataset, see Appendix~\ref{appendix:in_distribution}, while OOD evaluation on RESIDE-\(\beta\) (RTTS) and OTS datasets is detailed in Appendix~\ref{appendix:ood_performance}, demonstrating the pipeline’s robustness across diverse hazy conditions. Evaluation metrics include SSIM and PSNR for dehazing and mAP for object detection (Appendix~\ref{appendix:evaluation}). Figure~\ref{fig:dehazing} illustrates visibility improvement with AOD-Net, Figure~\ref{fig:labeled_dehazing_foggy} shows its impact on object detection after dehazing with AOD-NetX on Foggy Cityscapes, and Figure~\ref{fig:labeled_dehazing_rtts} presents performance on the RESIDE dataset.
\begin{table}[h]
\caption{Comparison of mean Average Precision (mAP) on clear and foggy conditions for different architecture variants. The performance change column quantifies the relative drop (red) or gain (green) in detection accuracy when transitioning from clear to foggy conditions.}
\label{tab:performance_drop}
\centering
\begin{center}
\begin{tabular}{lccc}
\toprule
\textbf{Architecture Variants} & \textbf{mAP (Clear)} & \textbf{mAP (Foggy)} & \textbf{Performance Change} \\
\toprule
YOLOv5x & 0.5644 & 0.4850 & \color{red}-14.07\%  \\ 
AOD-Net+YOLOv5x & 0.6813 & 0.5822 & \color{red}-14.53\%  \\ 
\textbf{YOLOv5s+AOD-NetX+YOLOv5x} & \textbf{0.4896} & \textbf{0.6152} & \color{green}+25.68\%  \\ 
YOLOv8x & 0.5243 & 0.4948 & \color{red}-5.63\% \\ 
AOD-Net+YOLOv8x & 0.6099 & 0.5900 & \color{red}-3.27\%  \\ 
\textbf{YOLOv8n+AOD-NetX+YOLOv8x} & \textbf{0.5150} & \textbf{0.6114} & \color{green}+18.71\% \\
 \bottomrule
\end{tabular}
\end{center}
\end{table}

An unexpected finding in our evaluation is the performance trend of models incorporating AOD-NetX. The proposed pipeline performed superior on foggy images and while conventional models like YOLOv5x, YOLOv8x and also their integration with AOD-Net show a natural drop in mAP when transitioning from clear to foggy conditions, architectures integrating AOD-NetX exhibit an inverse trend—performing better under foggy conditions than in clear ones (see Table~\ref{tab:performance_drop}). Notably, \textbf{YOLOv5s+AOD-NetX+YOLOv5x} achieves a 25.68\% relative mAP gain in foggy conditions, while \textbf{YOLOv8n+AOD-NetX+YOLOv8x} shows an 18.71\% relative gain, highlighting the anomalies we found in our experiments.

\section{Discussions}
\label{sec:discussion}
The unexpected performance improvement of AOD-NetX-based models under foggy conditions (Table~\ref{tab:performance_drop}) raises critical questions about feature adaptation in hybrid object detection pipelines. Typically, object detectors show a decrease in accuracy when transitioning from clear to foggy conditions, as seen in conventional YOLOv5x and YOLOv8x models. However, our proposed AOD-NetX integration leads to an inverse trend, where object detection improves under foggy conditions relative to that under clear conditions. This suggests that AOD-NetX introduces an implicit domain adaptation effect, which makes the detection network more attuned to foggy environments at the cost of generalization to clear images.

One possible explanation lies in bias and overprocessing in the data set. Since AOD-NetX is trained primarily on foggy images, its learned feature space is optimized for haze removal, but lacks the necessary constraints to preserve features in clear conditions. Consequently, when applied to clear images, the model introduces distortions instead of enhancements, disrupting feature consistency for the object detector. This emphasizes the need for context-aware enhancement, where image-processing techniques are selectively applied based on scene conditions rather than indiscriminately.

Furthermore, the results challenge the assumption that cascading pipelines, where lightweight detection informs region-specific enhancement before a final robust detection, always improve performance. While effective in foggy settings, this multi-stage approach appears to introduce trade-offs, potentially harming accuracy in clear conditions. Future designs must strike a balance between specialization for adverse weather conditions and adaptability to diverse environments. Another consideration is the real-time feasibility of this approach. RoI-specific dehazing adds computational overhead, which could limit deployment in time-sensitive applications such as autonomous driving. Optimizing processing efficiency while retaining performance gains remains an open challenge.
\section{Limitations \& Future Work}
\label{sec:limitations}
A fundamental limitation of integrating dehazing into object detection is the feature space misalignment between foggy and clear images. Models trained primarily on foggy conditions lack the ability to preserve the natural characteristics of clear images, leading to unintended alterations that degrade detection performance. This highlights the importance of adaptive enhancement techniques that can determine when dehazing is necessary, rather than applying it universally. A potential solution is the integration of a haze-level estimation module, which could prevent unnecessary processing by triggering dehazing only when haze exceeds a certain threshold~\citep{mao2014detecting}.

Another challenge is pretraining for scene differentiation. Since the AOD-NetX-enhanced models perform better in foggy conditions, their feature representations may be overfitting to haze-specific characteristics. Introducing joint training on both foggy and clear images could help mitigate this issue by aligning the feature space across different visibility conditions~\citep{huang2024degradation}.

Additionally, unifying dehazing and object detection into a single model rather than having multi-stage framework may yield mutual benefits. For instance, detection-aware dehazing—where dehazing prioritizes regions of interest—could help the model preserve essential features for object detection, enhancing accuracy in both clear and foggy conditions~\citep{fan2024friendnet}.

Finally, computational efficiency remains a key concern for real-time applications. While the current pipeline enhances detection performance in low-visibility conditions, its multi-stage nature introduces latency. Future work should focus on optimizing inference speed, exploring lightweight architectures, and developing efficient knowledge distillation techniques to maintain accuracy while reducing processing overhead.

By implementing \textbf{adaptive processing strategies}, \textbf{improved pretraining}, and \textbf{joint optimization}, future object detection pipelines can become more resilient across diverse visibility conditions, ensuring robust performance without compromising clarity in optimal conditions.

\bibliography{iclr2025}
\bibliographystyle{iclr2025}
\newpage
\appendix
\section{Datasets}
\label{appendix:datasets}

\subsection{Foggy CityScapes}
The Foggy Cityscapes dataset \citep{b31} was developed to tackle the challenge of semantic foggy scene understanding (SFSU). While significant research has been conducted on image dehazing and semantic scene understanding for clear-weather images, SFSU remains relatively underexplored. Due to the challenges associated with collecting and annotating real-world foggy images, synthetic fog is introduced into clear-weather outdoor scenes. This synthetic fog generation process utilizes incomplete depth information to simulate realistic foggy conditions on images from the Cityscapes dataset, resulting in a dataset comprising 20,550 images. The dataset is divided into a training set of 2,975 images, a validation set of 500 images, and a test set of 1,525 images. Key characteristics of the dataset include:

\begin{itemize}
\item \textbf{Synthetic Fog Generation}: Synthetic fog is added to real clear-weather images using a dedicated pipeline that incorporates the transmission map.
\item \textbf{Data Utilization}: The dataset supports both supervised and semi-supervised learning. A synthetic foggy dataset was generated using the synthetic transmission map, followed by supervised learning on the resulting foggy images.
\end{itemize}

\subsection{RESIDE-$\beta$}
The RESIDE-$\beta$ Outdoor Training Set (OTS) is a comprehensive dataset curated to support research in outdoor image dehazing. It addresses the degradation caused by haze in outdoor scenes, which negatively impacts image quality and downstream tasks such as object detection and semantic segmentation. The dataset contains approximately 72,135 outdoor images with varying haze intensities, allowing for robust training of dehazing algorithms. For evaluation, we use the RESIDE-\(\beta\) (REalistic Single Image DEhazing) dataset \citep{b32}. A subset of RESIDE-\(\beta\)
, the Real-Time Testing Set (RTTS), comprises 4,322 real-world hazy images with object detection annotations. The dataset is split into a training set of 3,000 images, a validation set of 500 images, and a test set of 1,500 images.
\section{Human Visual Cues}
\label{appendix:human_visual_cues}

\textbf{Selective Attention and Foveation}: The human eye does not perceive all areas of the visual field with equal clarity. Foveal vision, which corresponds to central vision, is highly detailed and is essential for tasks such as reading and object recognition. In contrast, peripheral vision is less detailed but more sensitive to motion. The visual system initially scans the entire scene using peripheral vision, akin to the preliminary detection phase in our approach. This broad scanning process helps identify regions requiring closer inspection, enabling a more detailed analysis through foveal vision. Similarly, the proposed method does not process every detail uniformly but prioritizes key areas of interest.

\textbf{Adaptation to Environmental Conditions}: The human visual system dynamically adjusts to varying lighting conditions and levels of visibility, such as adapting from bright sunlight to a dark room. Similarly, the adaptive dehazing method modulates its processing intensity and focus based on detection feedback and environmental context. This mechanism ensures optimal perception, mirroring the way human vision adapts to maintain clarity under diverse conditions.

\textbf{Eye Tracking and Gaze-Directed Processing}: Eye-tracking technology monitors gaze direction and identifies focal points of attention. This concept translates to strategically allocating resources toward regions of interest in computational visual processing. The proposed method follows a similar principle by directing dehazing and detailed object detection efforts to areas where objects are likely to be present. Just as human vision selectively fixates on specific regions when searching for an object, the system prioritizes certain parts of the image to enhance clarity and detection performance.

\textbf{Integration of Bottom-Up and Top-Down Processes}: Human vision combines bottom-up processing, driven by sensory input, with top-down processing, guided by prior knowledge, expectations, and goals. The proposed model adopts a similar dual approach: it first employs a bottom-up strategy, where object detection algorithms identify potential areas of interest. This is followed by a top-down refinement process, where dehazing efforts are concentrated on flagged areas, leveraging previous learning. This interplay between data-driven signals and cognitive insights aligns with the way human perception integrates sensory input with contextual understanding.
\section{Methodology}
\label{appendix:dehazing}

\textbf{Preliminary Detection}: A lightweight and fast object detection algorithm, such as YOLOv5s or YOLOv8n, is employed to rapidly scan the image and identify potential regions of interest or active regions. These models flag image patches with a high probability of containing objects. While the smaller variants of YOLO models offer lower accuracy compared to their full-sized counterparts, they are significantly faster, making them well-suited for this initial detection phase.

\textbf{Region-Based Dehazing}: Dehazing algorithms are selectively applied to the active regions identified during the preliminary detection phase. The approach dynamically adjusts based on the depth or severity of haze within the detected regions, ensuring an adaptive and efficient dehazing process.

\begin{figure}
    \centering
    \includegraphics[width=\linewidth]{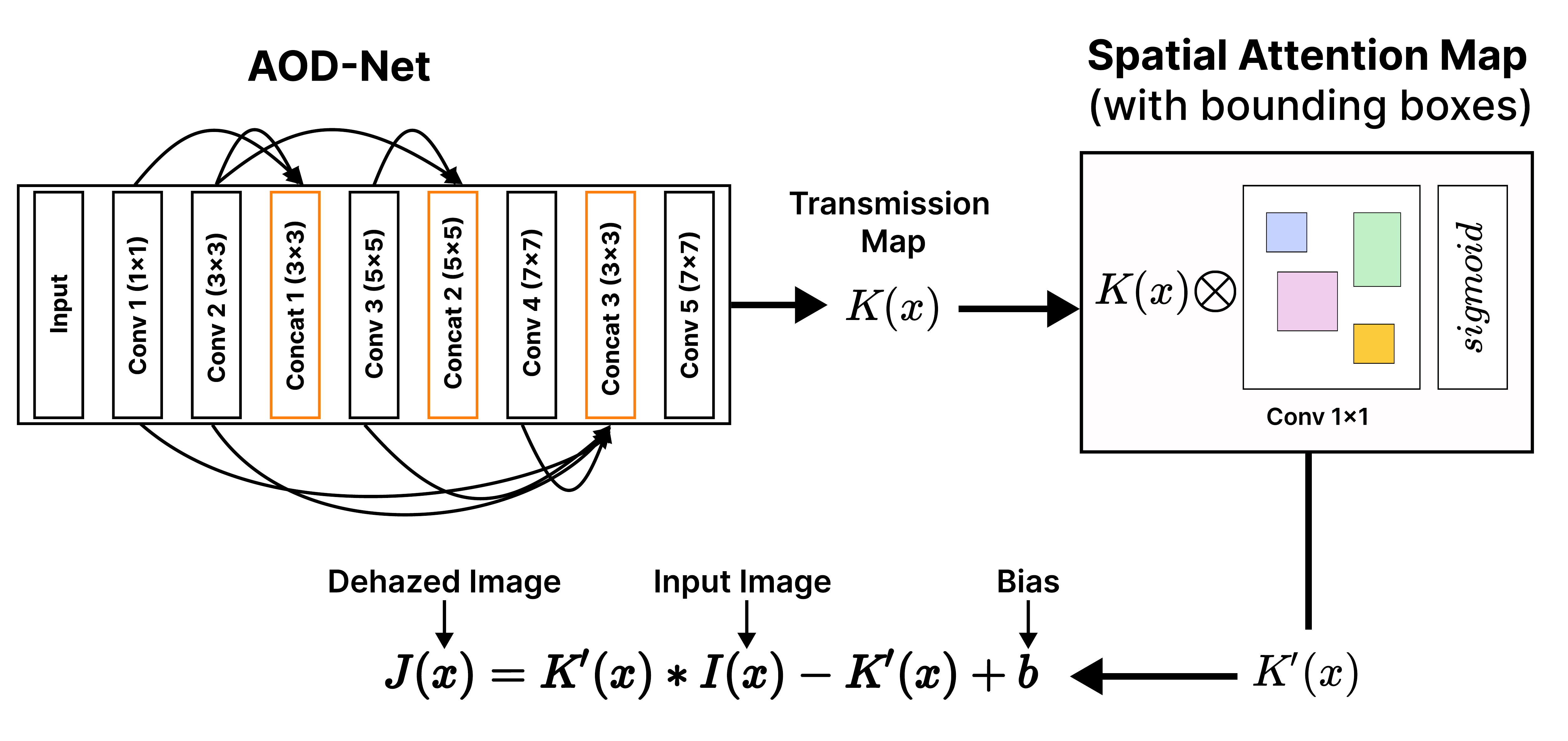}
    \caption{Architecture of AOD-NetX: The model takes the transmission map output, \( K(x) \), from AOD-Net and applies a spatial attention layer to emphasize key regions of interest (bounding boxes) in the input image. The refined transmission map, \( K'(x) \), is then utilized to dehaze the image.}
    \label{fig:aodnetx}
\end{figure}

The proposed architecture, \textbf{AOD-NetX}, illustrated in Figure \ref{fig:aodnetx}, builds upon the transmission map generated by the standard AOD-Net \citep{b34}. This transmission map is integrated into a spatial attention map module, producing an attention-enhanced transmission map. The spatial attention map is derived from the bounding boxes or Regions of Interest (ROIs) detected by the lightweight object detection model (YOLOv5s/YOLOv8n) within the proposed framework. A sigmoid layer is applied to map the output probabilities to a range between 0 and 1. Unlike softmax, which normalizes outputs across multiple regions, sigmoid is preferred in this context since each bounding box holds independent significance.
\section{Dehazing Models}
\label{appendix:dehazing_models}

\subsection{AOD-Net}
AOD-Net (All-in-One Dehazing Network) is a convolutional neural network (CNN) designed for haze removal by directly reconstructing the clean image in an end-to-end manner. Unlike traditional approaches that separately estimate transmission maps and atmospheric light, AOD-Net is based on a re-formulated atmospheric scattering model, allowing it to generate dehazed images without intermediate computations. This lightweight architecture delivers superior performance in terms of Peak Signal-to-Noise Ratio (PSNR) and Structural Similarity Index (SSIM) while also enhancing visual quality. Additionally, its modular design enables seamless integration into other deep learning models, such as Faster R-CNN, thereby improving object detection performance in hazy conditions.

\subsection{Dehaze-UNet}
The Dehaze-UNet model employs an encoder-decoder architecture with skip connections to effectively restore dehazed images. The encoder reduces spatial dimensions through successive convolutional layers, extracting essential features while maintaining structural integrity. Each convolutional layer is followed by batch normalization and ReLU activation, enhancing feature representation. The decoder then upscales these features, reintegrating spatial details from the encoder via skip connections, which are critical for preserving fine-grained textures lost during downsampling. The final stage consists of double convolution layers in the decoder for feature refinement, followed by Max Pooling and Bilinear Interpolation to optimize feature representation and ensure smooth transitions in the reconstructed image. This makes U-Net particularly effective for dehazing tasks.

\subsection{DehazeNet}
DehazeNet follows a structured four-stage approach, as illustrated in Figure 5 \citep{b35}. The first stage, feature extraction, employs 16 filters combined with four MaxOut layers, which use max pooling to reduce data dimensionality. The second stage, multi-scale mapping, captures both fine-grained and high-level features to ensure a comprehensive feature representation. The third stage, local extremum, utilizes a specialized max pooling operation to enhance spatial invariance while preserving image resolution. Finally, the non-linear regression stage incorporates Bilateral ReLU as the activation function, which constrains the output within a defined range to prevent oversaturation and maintain image clarity.

\subsection{AOD-NetX}
\label{appendix:aodnetX}
The proposed \textbf{AOD-NetX} architecture, depicted in Figure~\ref{fig:aodnetx}, extends AOD-Net by leveraging its transmission map within a spatial attention map module to generate an attention-focused transmission map. This spatial attention map is derived from the bounding boxes or Regions of Interest (ROIs) identified by the lightweight object detection model (YOLOv5s) within our framework. A sigmoid activation layer is applied to map the output probabilities to a range between 0 and 1. Unlike softmax, which normalizes outputs across multiple regions, sigmoid is preferred as each bounding box holds independent significance.

\section{Object Detection Models}
\label{appendix:object_detection}

The detection pipeline incorporates various YOLO models, each optimized for specific applications. YOLOv5s is a lightweight variant designed for real-time detection with minimal computational overhead, while YOLOv8n (Nano) is optimized for high-speed processing on resource-constrained devices, such as mobile phones. In contrast, YOLOv5x, with its CSP backbone and advanced data augmentation techniques, delivers enhanced performance for more complex scenes, whereas YOLOv8x (Extra Large) achieves maximum accuracy when handling large-scale datasets.

The detection workflow begins by applying YOLOv5s or YOLOv8n to foggy images to generate initial object annotations. These annotations, along with the original image, undergo dehazing using AOD-NetX. The resulting dehazed image is then processed with YOLOv5x or YOLOv8x, ensuring precise and refined detection outcomes.
\begin{table}[t]
\caption{Average Loss and SSIM scores of Dehazing techniques on Foggy Cityscapes dataset}
\begin{center}
    \begin{tabular}{lcc}
    \toprule
        \textbf{Dehazing Model} & \textbf{Average Loss} & \textbf{SSIM}\\
        \toprule
        AOD-Net & 0.0468 & 0.994 \\

          UNet-Dehaze & 0.0323 & 0.992 \\

         DehazeNet & 0.0572 & 0.991 \\
         \bottomrule
    \label{tab:dehzing_methods}
    \end{tabular}
    \end{center}
\end{table}
\begin{figure}[t]
    \centering
    \includegraphics[width=0.75\linewidth]{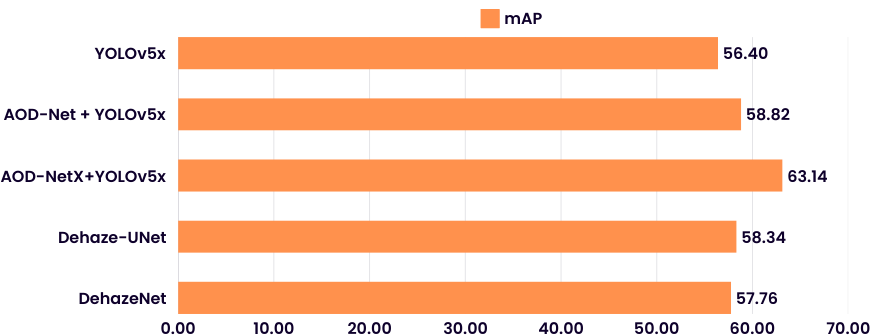}
    \caption{Comparison of mean Average Precision (mAP) for different dehazing and object detection module combinations.}
    \label{fig:map_compare}
\end{figure}
\section{Additional Results}
\label{appendix:additional_results}

The dehazing modules are trained independently on the provided datasets, while the object detection models (various YOLO versions) remain pre-trained on the MS-COCO dataset. This modular approach allows seamless integration of the dehazing module into existing detection pipelines without requiring full retraining. However, fine-tuning the entire architecture on target datasets could yield further performance improvements, making it a promising direction for future ablation studies.

Table \ref{dehazing} presents the comparative results, demonstrating that AOD-NetX generally outperforms the standard AOD-Net in terms of SSIM and PSNR across most datasets. For Foggy Cityscapes and RESIDE-$\beta$ OTS, AOD-NetX achieves higher SSIM and PSNR values, indicating superior structural similarity and signal fidelity. However, in the case of RESIDE-$\beta$ RTTS, while AOD-NetX attains a slightly higher PSNR, AOD-Net exhibits a significantly higher SSIM score, suggesting better structural detail retention in this specific dataset. Overall, AOD-NetX proves more effective in most scenarios, particularly under complex foggy conditions.
\begin{table}[t]
\caption{Performance of dehazing methods: AOD-Net and AOD-NetX}
\label{dehazing}
\begin{center}
\begin{tabular}{llccc}
\toprule
\multicolumn{1}{c}{\bf Dataset} & \multicolumn{1}{c}{\bf Dehazing Method} & \multicolumn{2}{c}{\bf Evaluation Metrics} \\
\multicolumn{1}{c}{} & \multicolumn{1}{c}{} & \multicolumn{1}{c}{\bf SSIM} & \multicolumn{1}{c}{\bf PSNR} \\
\toprule
\multirow{2}{*}{\bf Foggy Cityscapes} & AOD-Net & 0.994 & 26.74 \\
                                  & AOD-NetX     & \textbf{0.998} & \textbf{27.22} \\
\multirow{2}{*}{\bf RESIDE-$\beta$ OTS} & AOD-Net & 0.920 & 24.14 \\
                                    & AOD-NetX     & \textbf{0.945} & \textbf{25.80} \\
\multirow{2}{*}{\bf RESIDE-$\beta$ RTTS} & AOD-Net & \textbf{0.932} & 27.59 \\
                                     & AOD-NetX     & 0.656 & \textbf{27.62} \\
\bottomrule                                     
\end{tabular}
\end{center}
\end{table}
\subsection{In-Distribution Performance of Perceptual Piercing}
\label{appendix:in_distribution}

The evaluation results of Perceptual Piercing variations, trained and tested on the Foggy Cityscapes dataset, are presented in Table~\ref{IN}. The integration of dehazing modules, such as AOD-Net and AOD-NetX (detailed in Appendix~\ref{appendix:dehazing}), consistently enhances object detection in both clear and foggy conditions.

Among the tested variants, the AOD-Net + YOLOv5x configuration achieved the highest mAP under clear conditions (0.6813). In foggy conditions, YOLOv5s + AOD-NetX + YOLOv5x and YOLOv8n + AOD-NetX + YOLOv8x demonstrated the best performance, with mAP scores of 0.6152 and 0.6114, respectively. In contrast, baseline YOLO models (YOLOv5x and YOLOv8x) exhibited lower detection accuracy, highlighting the effectiveness of advanced dehazing techniques in low-visibility environments.
\begin{table}[t]
\caption{\textbf{Train}- Foggy Cityscapes, \textbf{Test}- Foggy Cityscapes: Evaluation of various Perceptual Piercing variations based on mean Average Precision (mAP) under both clear and foggy conditions.}
\label{IN}
\begin{center}
\begin{tabular}{ccc}
\toprule
\multicolumn{1}{c}{\bf Architecture Variants} & \multicolumn{1}{c}{\bf Conditions} & \multicolumn{1}{c}{\bf Evaluation Metrics (mAP)} \\
\toprule
\multirow{2}{*}{\bf YOLOv5x} & Clear & 0.5644 \\
                                  & Foggy     & 0.485\\
\multirow{2}{*}{\bf AOD-Net+YOLOv5x} & Clear & 0.6813 \\
                                  & Foggy     & 0.5822\\
\multirow{2}{*}{\bf YOLOv5s+AOD-NetX+YOLOv5x} & Clear & 0.4896 \\
                                  & Foggy     & 0.6152\\
\multirow{2}{*}{\bf YOLOv8x} & Clear & 0.5243 \\
                                  & Foggy     & 0.4948\\
\multirow{2}{*}{\bf AOD-Net+YOLOv8x} & Clear & 0.6099 \\
                                  & Foggy     & 0.5900\\
\multirow{2}{*}{\bf YOLOv8n+AOD-NetX+YOLOv8x} & Clear & 0.5150 \\
                                  & Foggy     & 0.6114\\

                                  \bottomrule
\end{tabular}
\end{center}
\end{table}
\subsection{Out-of-Distribution Performance of Perceptual Piercing}
\label{appendix:ood_performance}

The evaluation results in Table~\ref{OOD}, where Perceptual Piercing variations were trained on Foggy Cityscapes and tested on the RESIDE-$\beta$ OTS and RTTS datasets, highlight key performance trends. The YOLOv8x architecture achieved the highest mAP scores under foggy conditions, with 0.7125 on OTS and 0.6978 on RTTS. Among the YOLOv5 variants, the baseline YOLOv5x model performed best, achieving 0.6944 on OTS and 0.6655 on RTTS.

The addition of AOD-Net generally enhanced performance for YOLOv8 but had a diminishing effect on YOLOv5. Meanwhile, models incorporating AOD-NetX exhibited lower mAP values across both test datasets, suggesting that its integration may require further optimization. Overall, the results indicate that YOLOv8x is more robust in handling foggy conditions compared to other model variations.
\begin{table}[t]
\caption{\textbf{Train}- Foggy Cityscapes, \textbf{Test}- RESIDE-$\beta$ OTS and RTTS: Evaluation of various Perceptual Piercing variations based on mean Average Precision (mAP) under foggy conditions.}
\label{OOD}
\begin{center}
\begin{tabular}{ccc}
\toprule
\multicolumn{1}{c}{\bf Architecture Variants} & \multicolumn{1}{c}{\bf Configuration} & \multicolumn{1}{c}{\bf Evaluation Metrics (mAP)} \\
\toprule
\multirow{2}{*}{\bf YOLOv5x} & Test: OTS & 0.6944 \\
                                  & Test: RTTS     & 0.6655\\
\multirow{2}{*}{\bf AOD-Net+YOLOv5x} &Test: OTS & 0.6325 \\
                                  & Test: RTTS     & 0.6156\\
\multirow{2}{*}{\bf YOLOv5s+AOD-NetX+YOLOv5x} & Test: OTS & 0.5679 \\
                                  & Test: RTTS    & 0.5297\\
\multirow{2}{*}{\bf YOLOv8x} & Test: OTS & 0.7125 \\
                                  & Test: RTTS    & 0.6978\\
\multirow{2}{*}{\bf AOD-Net+YOLOv8x} & Test: OTS & 0.6458 \\
                                  & Test: RTTS     & 0.6125\\
\multirow{2}{*}{\bf YOLOv8n+AOD-NetX+YOLOv8x} & Test: OTS & 0.5779 \\
                                  & Test: RTTS    & 0.5312\\
                                  \bottomrule
\end{tabular}
\end{center}
\end{table}
\begin{figure}[ht]
    \centering
    \subfigure[Foggy Cityscapes: Before Dehazing]{\fbox{\includegraphics[width=0.45\textwidth]{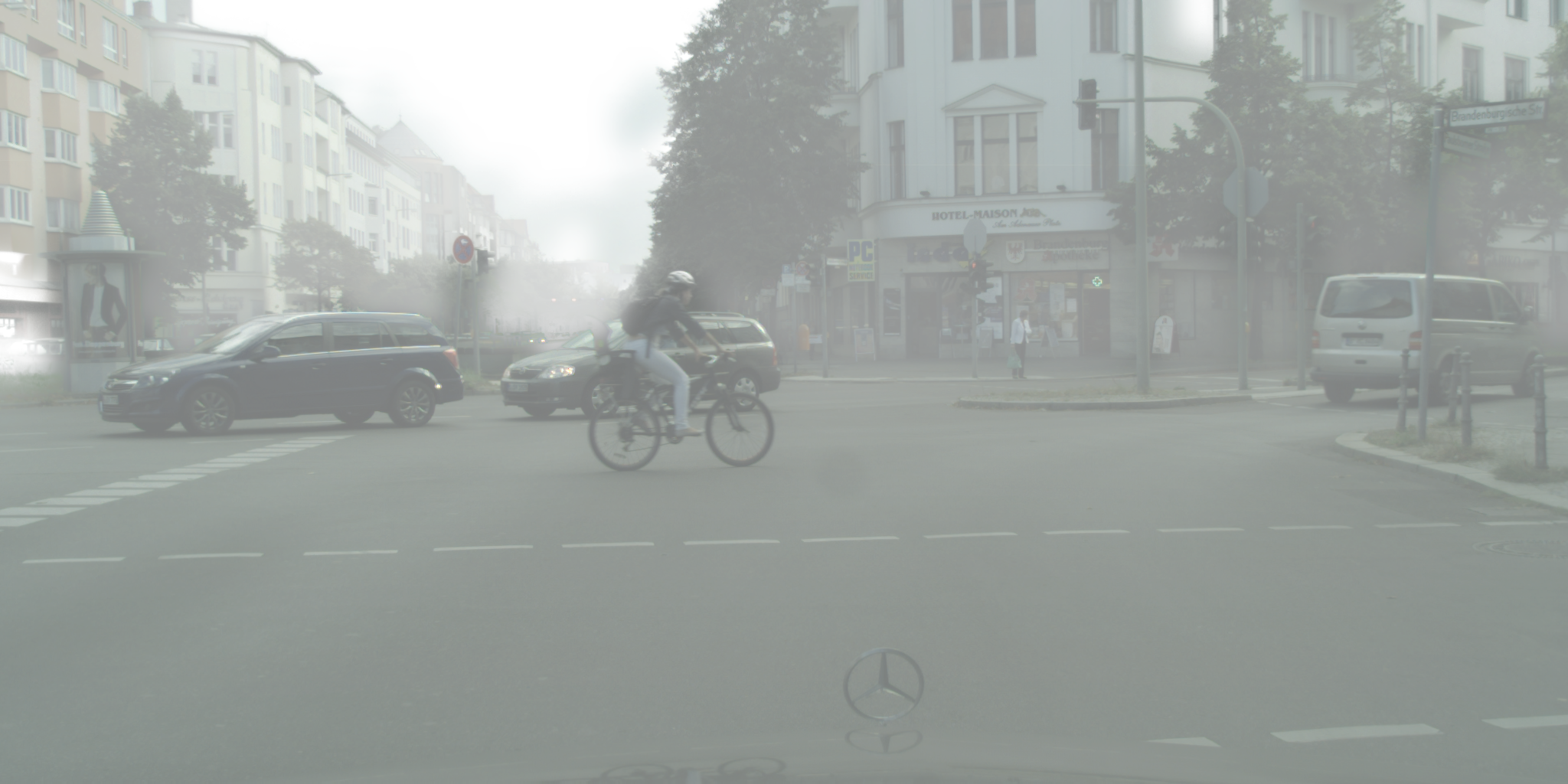}}}\hfill
    \subfigure[Foggy Cityscapes: After Dehazing (using AOD-NetX)]{\fbox{\includegraphics[width=0.45\textwidth]{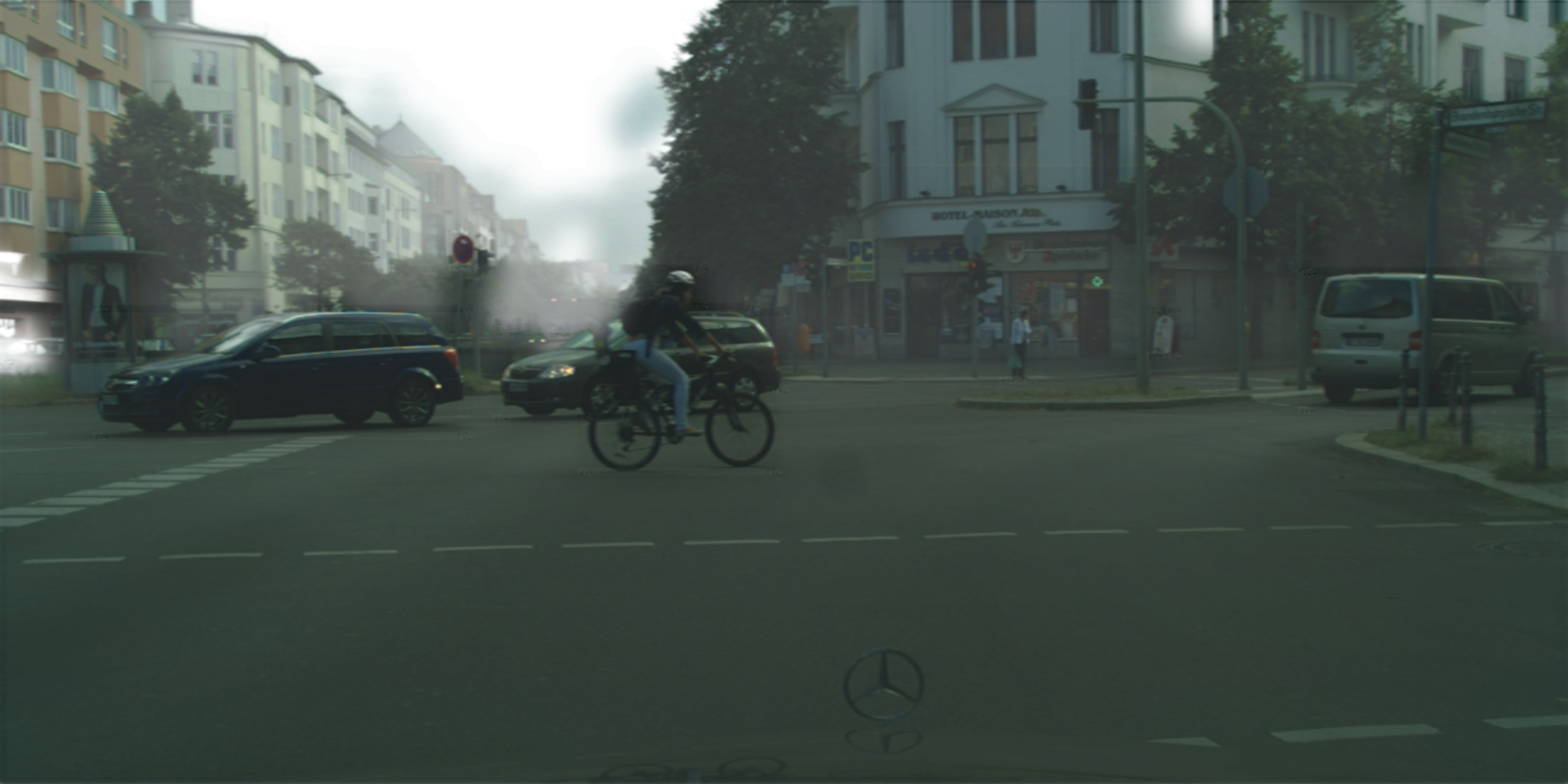}}}
    \caption{Dehazing performance on Foggy Cityscapes dataset.}
    \label{fig:dehazing}
\end{figure}
\begin{figure}[ht]
    \centering
    \subfigure[Foggy Cityscapes: Before Dehazing]{\fbox{\includegraphics[width=0.45\textwidth]{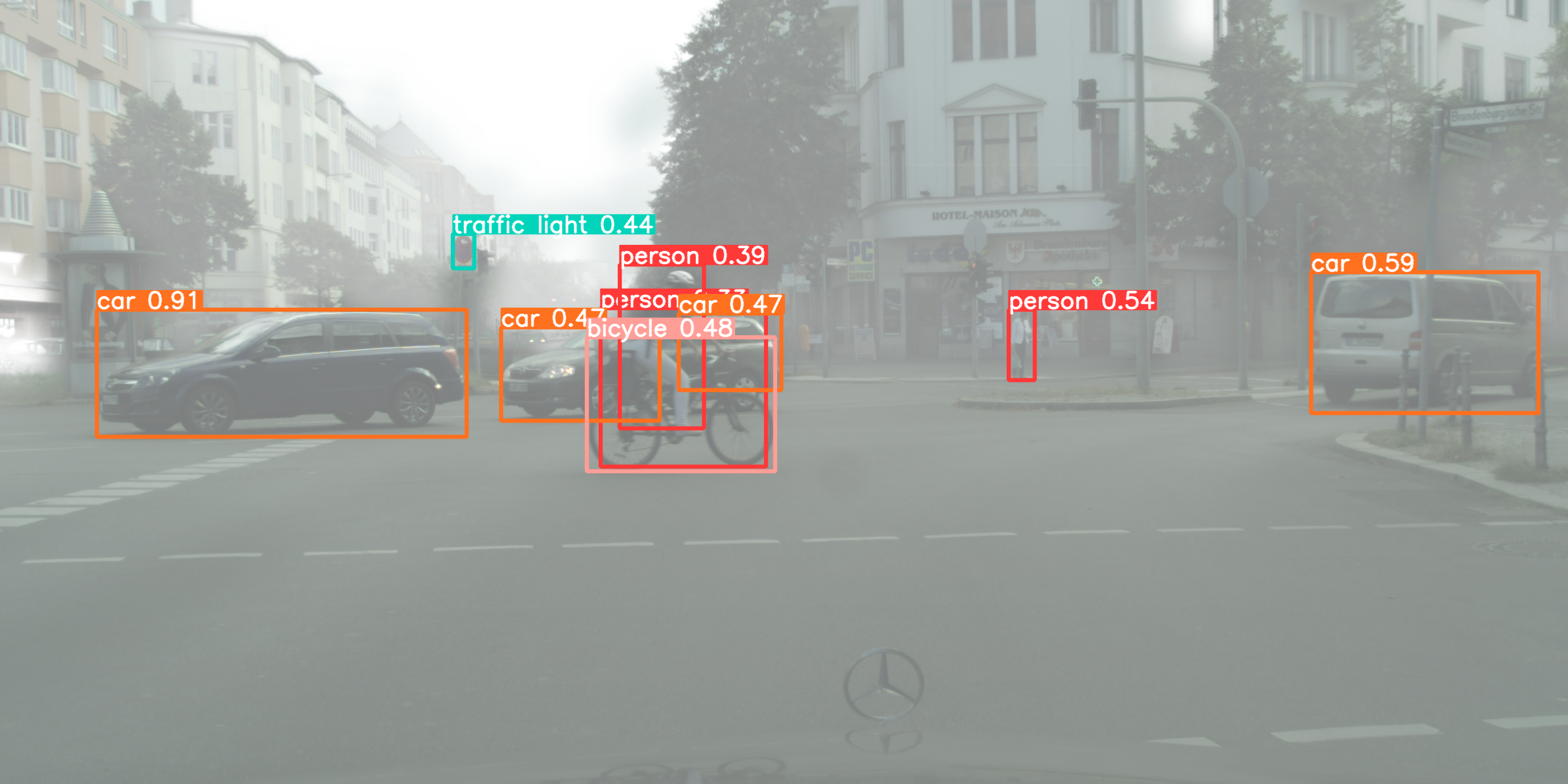}}}\hfill
    \subfigure[Foggy Cityscapes: After Dehazing (using AOD-NetX)]{\fbox{\includegraphics[width=0.45\textwidth]{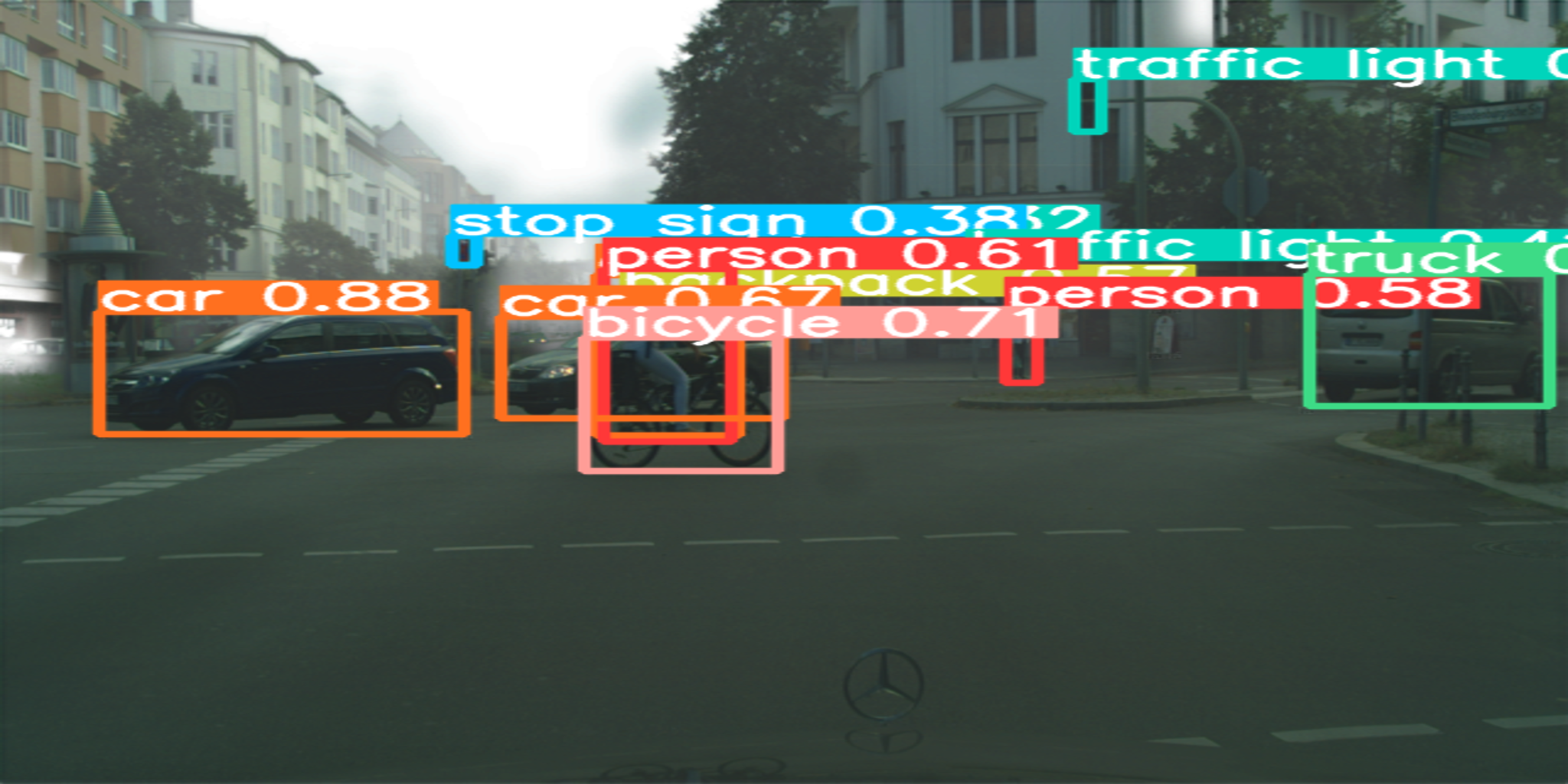}}}
    \caption{Dehazing performance on Foggy Cityscapes dataset.}
    \label{fig:labeled_dehazing_foggy}
\end{figure}
\begin{figure}[ht]
    \centering
    \subfigure[RESIDE-$\beta$: Before Dehazing]{\fbox{\includegraphics[width=0.45\textwidth]{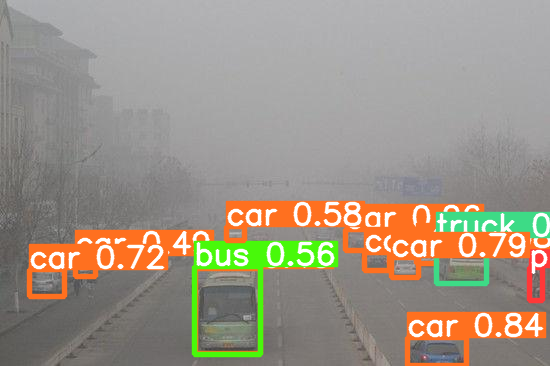}}}\hfill
    \subfigure[RESIDE-$\beta$: After Dehazing (using AOD-NetX)]{\fbox{\includegraphics[width=0.45\textwidth]{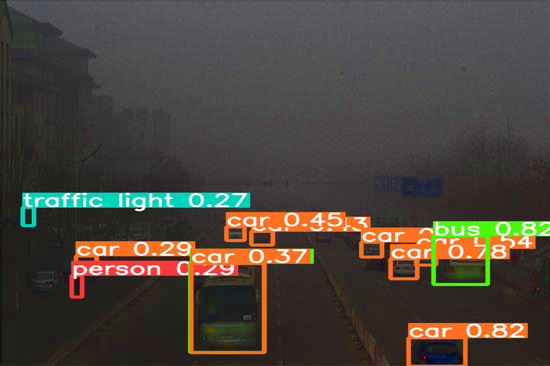}}}
    \caption{Dehazing performance on RESIDE-$\beta$ dataset.}
    \label{fig:labeled_dehazing_rtts}
\end{figure}
\section{Evaluation Metrics}
\label{appendix:evaluation}

\subsection{Structural Similarity Index Measure (SSIM)}
\label{ssim}
The performance of dehazing methods is evaluated using the Structural Similarity Index Measure (SSIM), which quantifies the similarity between two images based on luminance, contrast, and structural components. It is defined as:

\begin{equation}
    SSIM(x, y) = \frac{(2 \mu_x \mu_y + c_1)(2 \sigma_{xy} + c_2)}{(\mu_x^2 + \mu_y^2 + c_1)(\sigma_x^2 + \sigma_y^2 + c_2)}
\end{equation}

where:
\begin{itemize}
    \item $\mu_x$ and $\mu_y$ are the mean intensities of images $x$ and $y$, respectively.
    \item $\sigma_x^2$ and $\sigma_y^2$ denote the variances of $x$ and $y$.
    \item $\sigma_{xy}$ represents the covariance between $x$ and $y$.
    \item $c_1 = (k_1 L)^2$ and $c_2 = (k_2 L)^2$ are stabilizing constants to prevent division by zero, where $L$ is the dynamic range of pixel values (e.g., 255 for 8-bit images), and default values are $k_1=0.01$ and $k_2=0.03$.
\end{itemize}

\subsection{Peak Signal-to-Noise Ratio (PSNR)}
\label{psnr}
Peak Signal-to-Noise Ratio (PSNR) is a widely used metric to assess image reconstruction quality by comparing the original and processed images. It is expressed in decibels (dB) and is calculated as:

\begin{equation}
\text{PSNR} = 10 \cdot \log_{10} \left( \frac{\text{MAX}^2}{\text{MSE}} \right),
\end{equation}

where $\text{MAX}$ is the maximum possible pixel value (e.g., 255 for 8-bit images), and $\text{MSE}$ is the Mean Squared Error:

\begin{equation}
\text{MSE} = \frac{1}{mn} \sum_{i=0}^{m-1} \sum_{j=0}^{n-1} \left( I(i,j) - K(i,j) \right)^2.
\end{equation}

Here, $I(i,j)$ and $K(i,j)$ represent pixel values at position $(i,j)$ in the original and reconstructed images, respectively. A higher PSNR value indicates better image quality, as it corresponds to lower distortion. PSNR is extensively used in evaluating dehazing, denoising, and image compression methods.

\subsection{Mean Average Precision (mAP)}
\label{map}
For object detection performance, we use mean Average Precision (mAP), which evaluates the precision-recall tradeoff. The Average Precision (AP) is computed as:

\begin{equation}
    AP = \frac{\sum_{k=1}^{n} (P(k) \times \text{rel}(k))}{\text{number of relevant objects}}
\end{equation}

where:
\begin{itemize}
    \item $P(k)$ is the precision at rank $k$.
    \item $\text{rel}(k)$ is an indicator function, which is 1 if the object at rank $k$ is relevant, and 0 otherwise.
    \item $n$ is the total number of retrieved objects.
\end{itemize}

The mean Average Precision is computed as:

\begin{equation}
    mAP = \frac{\sum_{q=1}^{Q} AP_q}{Q}
\end{equation}

where $AP_q$ is the Average Precision for the $q^{th}$ query, and $Q$ is the total number of queries. Higher mAP values indicate better object detection performance across different classes.

\end{document}